\documentclass[11pt,a4paper]{article}
\usepackage[hyperref]{acl2021}
\usepackage{times}
\usepackage{latexsym}

\usepackage{microtype}
\usepackage{booktabs}
\usepackage{multirow}
\usepackage{makecell}
\usepackage{graphicx}

\aclfinalcopy 

\title{How Good is Automatic Segmentation as a Multimodal Discourse Annotation Aid?}

\author{Corbyn Terpstra, Ibrahim Khebour, Mariah Bradford, Brett Wisniewski, \\ {\bf Nikhil Krishnaswamy and Nathaniel Blanchard}  \\
  Department of Computer Science \\
  Colorado State University \\
  Fort Collins, CO, USA \\
  \texttt{\{cytoniv,nblancha\}@colostate.edu}}

\date{}

\begin{document}
\maketitle
\begin{abstract}

Collaborative problem solving (CPS) in teams is tightly coupled with the creation of shared meaning between participants in a situated, collaborative task. In this work, we assess the quality of different utterance segmentation techniques as an aid in annotating CPS. We (1) manually transcribe utterances in a dataset of triads collaboratively solving a problem involving dialogue and physical object manipulation, (2) annotate collaborative moves according to these gold-standard transcripts, and then (3) apply these annotations to utterances that have been automatically segmented using toolkits from Google and OpenAI's Whisper. We show that the oracle utterances have minimal correspondence to automatically segmented speech, and that automatically segmented speech using different segmentation methods is also inconsistent. We also show that annotating automatically segmented speech has distinct implications compared with annotating oracle utterances---since most annotation schemes are designed for oracle cases, when annotating automatically-segmented utterances, annotators must invoke other information to make arbitrary judgments which other annotators may not replicate. We conclude with a discussion of how future annotation specs can account for these needs. 
\end{abstract}

\section{Introduction}
\vspace*{-2mm}

In order for Artificially Intelligent (AI) agents to interact with with an environment, they must first accurately perceive that environment. In real-world contexts, this necessitates automatically preprocessing various modalities for downstream procedures. For example, an AI agent to modulate classroom discourse needs to first identify distinct discourse components, but a single spoken utterance from a team member could contain multiple discourse components. The identification of each discourse component within the utterance could easily spiral into a doctoral thesis but overly fixating on this preprocessing step would make it extremely difficult to make substantive progress on AI agents themselves. 

Typically, researchers default to ``oracle" data, where one assumes the preprocessing step has been completed with human level accuracy (e.g., human transcriptions of speech, utterances segmented by dialogue move). However, in a real-world agent deployment, preprocessing of data that would be fed into the automated system will instead be handled by off-the-shelf software.
Current practice in AI assumes the existence of suitable datasets that contain examples of the information an automated system is intended to extract. The task of developing the AI model entails solving for the function that best maps from the input samples to the desired outputs. If these datasets do not already exist, then the information that is to be learned must be annotated by humans.

Consider the scenario we focus on in this paper: a group collaborating to solve a problem involving the shared manipulation of physical objects.  Multiple modalities are implicated in such a task---group members speak to each other, but also point or gesture, use body language, and manipulate objects to communicate meaning and intent.  Specs intended for annotating {\it collaborative problem solving} (CPS) skills on display are intended to be used at the utterance level, and assume that the utterance has been segmented and transcribed by humans (``oracles'').  There are many frameworks for modeling CPS that have been developed by researchers in the learning sciences (e.g.,~\citet{roschelle1995construction,cukurova2018nispi,andrews2020exploring,sun_towards_2020}) and this literature stresses the multimodal nature of CPS~\cite{dillenbourg2006sharing}. For example, the occurrence of an interruption or the content of cross-talk may not be immediately evident from the audio signal alone, but watching the speakers interact may make it clear who is speaking when or what is said.  High-quality annotation of oracle utterances of a multimodal task like CPS therefore relies on annotators attending to the multiple modalities implicated while making their decisions. If the annotations are performed without this information, or with this information scrambled somehow, we should expect this to affect the quality of the annotation. The question is, how much?


The development of such annotation schemes is typically conducted separately from the rapid preprocessing and scaling that AI practitioners are likely to encounter when they use such annotated data for model training. There may be little that AI practitioners can ask of spec developers given the risk involved with the development of an annotation spec (one can imagine the truly unpleasant experience of developing an annotation spec and finding, after innumerable modeling and annotation cycles, that meaning captured in the spec is not linked to the expected meaningful outcomes). Further, spec developers may (arguably rightly) say that they have no expectation that their spec will be used to train AI models, and that the problems that unfold should be solved by AI developers themselves. These are quite reasonable arguments---and unless the annotation spec is being explicitly developed for AI systems, annotators are unlikely to change (we strongly encourage annotations developed for AI to think deeply about these problems---but such thoughts are outside of the scope of this particular paper). Nonetheless, AI development relies on interoperable annotated data, and as AI practitioners ourselves, we conclude that AI practitioners must think deeply about traditional annotation schemes and how we can best accommodate them.


In this paper, the annotation scheme we use is the one developed by \citet{sun_towards_2020} but the problem we address is independent of any particular spec.  Namely, when an annotation spec designed for one utterance segmentation method is applied to utterances automatically segmented using a different method, the information retrieved is different from what the original spec intended to encode.

We annotate utterances for CPS using expert annotators, and we also have expert annotators label, as best they can, automatically-segmented utterances. We discuss common strategies to transfer oracle annotations to real-world annotations. We underscore, exactly, how disconnected the oracle utterance labels may be from the labels on automatically-segmented utterances. Finally, we discuss how, given even just two automatic utterance segmentation methods, achieving a gold-standard annotating become quickly intractable if the specification itself does not contain strategies for accommodating suboptimal preprocessing. 

\vspace*{-2mm}
\section{Related Work}
\vspace*{-2mm}

The gap between oracle data and real-world data has been identified previously~\cite{blanchard2016semi}. Other works have pointed out the need to move away from oracle transcriptions in pursuit of AI applications for real-world use cases \cite{morbini2013asr,blanchard2018getting}. The use of automatic segmentation of speech for modeling tasks is becoming increasingly widespread \cite{bradford2022deep,bradford2022challenges,castillon2022multimodal}. 

Modeling in general has become more aware of the needs of real-world systems. For example, methods for automatically detecting mind wandering have moved from balanced datasets to heavily imbalanced datasets in acknowledgement of the need for such models to operate in the context of real-world distributions \cite{kuvar2022automatically}.

What is distinct with this work is that here we focus our analysis on the annotation implications, rather than on attempts to fix issues that arise through machine learning directly. For example, \citet{blanchard2018getting} refused to use human transcriptions in a multimodal sentiment challenge because such transcripts were not true to real-world contexts; however, they did not comment on how the labeling of sentiment might change were those annotations done on automatically extracted data.

Here, we explicitly focus on that challenge.  We explore the implications of segmentation and transcription methods when annotating CPS for groups. CPS is a critical skill used in many areas of life \cite{graesser_advancing_2018}, and AI agents for group settings will need some way of representing group state. Work has been done to model CPS at the utterance level~\cite{stewart_multimodal_2021,bradford2023automatic}. The framework defined by~\citet{sun_towards_2020} captures CPS at three levels and identifies specific actions that indicate different types of collaborative actions and their impact on group state. In particular, we hope our efforts here facilitate consistency across future CPS modeling efforts and meaningfully contribute to the CPS framework defined by \citet{sun_towards_2020}, and in general, we hope to prompt thought about annotation spec design and strategy in the face of potential uses involving automated preprocessing.

\vspace*{-2mm}
\section{Dataset}
\vspace*{-2mm}

Our dataset consists of audiovisual recordings of 10 triads performing a shared collaborative task which was developed to promote rich collaboration via multimodal communication. The task is performed by triads at a round table in a laboratory setting. The equipment on the table includes 6 blocks (of varying weight, size, and color), a balance scale, a worksheet demarcated with spaces to place the blocks (indicated with weights in grams), and a laptop on which participants submit their responses to survey questions throughout the task.

Participants are first given a balance scale to determine the weights of five of the colored wooden blocks. They are told that one block weighs 10 grams, but that they have to determine the weights of the rest of the blocks using the balance scale.\footnote{The pattern to the weights of the blocks is based on the Fibonacci sequence.} As the weight of each block is determined, participants place it on the worksheet next to its corresponding weight. The participants also must submit their final answer for the weight of each block to the survey form on the laptop. Once the weights of all five blocks are solved for, participants are given the sixth block and must identifying its weight \textit{without} using the scale (i.e., participants have to deduce the weight based on the pattern observed in the initial block weights). Finally, participants are asked to determine the weight of another mystery block that is \textit{not physically present} and explain how they arrived at the answer. The participants once again submit their answer as a group in the online survey and are given two chances (with a hint after the first guess if it is incorrect).

The total dataset consists of 10 videos, containing 3 participants each, for a total of 170 minutes of video. Participants ranged from 19--35 years old, recruited from a university population. 20\% were female while 80\% were male. 60\% were Caucasian non-Hispanic, 10\% were Hispanic/Latino, and 30\% were Asian. All volunteers spoke English through the task but spoke a variety of native languages.

Although this data was collected in a lab, the complexity of human-human interaction is appropriately captured in these recordings --- participants talk over each other, they speak with disfluencies, they interrupt each other, they engage in long run-on sentences punctuated by only a single em-dash, and they pause in the middle of sentences before resuming their thought. All of these complications make utterance segmentation quite difficult, and often these ambiguities are only resolved by human annotators with recourse to the visual modality.

\vspace*{-2mm}
\section{Preprocessing}
\vspace*{-2mm}



\subsection{Automatic Segmentation of Speech}
\vspace*{-1mm}

Automatic Speech Recognition, or ASR, approaches, must necessarily determine the boundaries of utterances.  Each ASR model segments audio in unique ways. This can be either through waiting for any pause in the audio, or waiting until a break of a certain length is encountered. ASR allows for AI to break apart the speech for the listener to in principle break down the amount of empty noise within audio recordings, and different systems using the same ASR component are interoperable on this level.

\vspace*{-2mm}
\subsection{Whisper}
\vspace*{-1mm}

Whisper \cite{radford2022robust} is a speech recognition system developed by OpenAI that was trained on 680,000 hours of audio to accurately determine and transcribe speech across many different languages. Whisper takes audio files and will listen to the first 30 seconds, or less depending on the length of the file, to determine the language of the speech. It will then segment the audio into full second segments, and will rarely cut off before a single or multiple full seconds have passed. Whisper is also optimized to segment audio into full sentences instead of simply looking for a break in the audio. In principle, this allows for transcription of long audio segments (e.g., lectures or speeches) with a fidelity closer to human transcription.

\vspace*{-2mm}
\subsection{Google ASR}
\vspace*{-1mm}

Google ASR \cite{velikovich2018semantic} is a speech recognition system released by Google, Inc. Google ASR listens for what it assumes to be human speech and attempts to transcribe what it hears. Google also will attempt to segment audio wherever it finds a break in speech. If a word is not picked up by the microphone correctly or is slightly inaudible, then Google will cut off the word and move on with the next segmentation. This could mean cutting off a thought mid-sentence, or removing words entirely from what someone is saying.

\vspace*{-2mm}
\section{Annotation Methodology}
\vspace*{-2mm}

Videos were first hand-transcribed to ensure the accuracy of the transcriptions. These hand, or oracle transcriptions, were then measured against the transcriptions from both Google ASR and Whisper. 

\vspace*{-2mm}
\subsection{Annotation Procedure}
\vspace*{-1mm}

When annotating the oracle files, speech start and end times would be marked down to the hundredth of a second. Each audio file would then be segmented into proper sentences or thoughts if the sentences were not completed. If people within the audio recording spoke over each other, each person's sentence was recorded as closely as possible, each with its own beginning and ending timestamp. This was done for each audio file from the 10 separate groups. Each segmented utterance was then coded by expert annotators using an updated version of the framework developed by~\citet{sun_towards_2020}. The annotators initially annotated all 10 videos separately, to get familiar with the framework, then were trained by experts in the framework on one video, while discussing how each CPS indicator would align with the weights task. The experts then annotated another video with a Fleiss' kappa score of 0.62 (agreement over 96\% of the number of subjects to be coded). 

\vspace*{-2mm}
\section{Transferring Annotations from Oracle to Automatically-Segmented Utterances}
\vspace*{-2mm}

Once oracle utterances are labeled, we map those labels to the automatically segmented utterances. The approach for that mapping depends on the task at hand and the type of labels we see. In the case of labeling collaborative problem solving (CPS), the multiclass binary labels can be inherited from the oracle segments to the automatic segments using overlap in timestamps. This is because the labels all still exist during that period. However, we lose label accuracy when we lose the exact timestamp where the label occurred. Another option is to only apply labels that occur in every oracle included in the segment; however, with CPS, this would rarely occur and we would lose most of our labels. 

\vspace*{-2mm}
\section{Effects of Oracle vs. Automatic Utterance Segmentation}
\vspace*{-2mm}


\paragraph{Count of utterances}
Table~\ref{tab:auto-oracle} shows the different number of utterances segmented out by each method for each group.

\begin{table}[h!]
\centering
\addtolength{\tabcolsep}{-3.5pt}  
\begin{tabular}{lcccccccccl}\toprule
\small Group &
  \small  1 &
  \small  2 &
  \small  3 &
  \small  4 &
  \small  5 &
  \small  6 &
  \small  7 &
  \small  8 &
  \small  9 &
  \small  10 \\ \midrule
  \small Whisper & \small 297 & \small 201 & \small 391 & \small 293 & \small 406  & \small 278 & \small 311 & \small 354 & \small 136 & \small 346 \\
  \small Google & \small 139 & \small 151 & \small 254 & \small 128 & \small 146  & \small 153 & \small 380 & \small 235 & \small 90 & \small 146 \\   \small Oracle &\small 229 &\small 207 &\small 337 &\small 195 &\small 237  & \small 227 &\small 590 &\small 338 &\small 134 & \small 379 \\ 
\bottomrule
\end{tabular}
\vspace*{-2mm}
\caption{\# of utterances per group determined by each segmentation method. Totals: Whisper - 3,013 utterances; Google - 1,822 utterances; Oracle - 2,873.}
\label{tab:auto-oracle}
\vspace*{-2mm}
\end{table}

Almost uniformly, Whisper segments more individual utterances than occur in the oracle transcripts, due to breaking up single oracle utterances into multiples (exceptions are groups 7 and 10).
Across all groups, Google segmentation creates fewer (sometimes far fewer) utterances than exist in the oracle, due to dropping utterances entirely or mistaking speech for background noise.  Google ASR does perform very well removing empty space from audio files compared to Whisper. 

\paragraph{Intrinsic ASR metrics}

Evaluation of the automatic speech transcription itself after automatic segmentation can be used as a proxy for information lost in part due to the segmentation process. Since error rates must be calculated with respect to the same set of utterances in order to be directly comparable, we focused this analysis on the transcription of Google-segmented utterances. Given an oracle transcript with assumed insertion, deletion, substitution, and total word error rates of 0, we observe that while overall word error rate (WER) is similar using Google and Whisper (Google: 0.573; Whisper: 0.542), Google has higher rates of substitutions (words in the oracle swapped for a different word) and deletions (words in the oracle removed by automated transcription), while Whisper has a significantly higher rate of insertions (words in the automated transcript not in the oracle). See Table~\ref{table_wer}.  

We investigated why Whisper had far more insertions and found it was linked to Google utterances that did not contain any speech. Occasionally, when listening to the audio files Google will hear empty noise as speech and create a segment for it. 
When feeding Whisper an audio segment containing only background noise, it would generate its own sentence to fill the void, and would occasionally choose a random language to generate the utterance in as well. This does not pose an issue in most situations, since the main purpose of ASR and transcription software would be to transcribe and recognize actual speech in audio files. Thus, the WER of Whisper seems to be partially be a product of our decision to use Google utterances. An appropriate method to filter out such segments, or, the use of Whisper's own segmentation would likely substantially lower the WER of whisper.

\begin{table*}[]
\centering
\begin{tabular}{lcccccccc}\toprule
        & \multicolumn{4}{c}{\small \textbf{Google}} & \multicolumn{4}{c}{\small \textbf{Whisper}} \\ \cmidrule(lr){2-5}\cmidrule(lr){6-9}
\small Group   & \small WER & \small Sub. rate & \small Del. rate & \small Ins. rate & \small WER & \small Sub. rate & \small Del. rate & \small Ins. rate \\ \hline
\small 1 & \small 0.571 & \small 0.252 & \small 0.113 & \small 0.206 & \small 0.534 & \small 0.193 & \small 0.045 & \small 0.296     \\
\small 2 & \small 0.459 & \small 0.211 & \small 0.128 & \small 0.120 & \small 0.416 & \small 0.177 & \small 0.040 & \small 0.200     \\
\small 3 & \small 0.539 & \small 0.236 & \small 0.117 & \small 0.186 & \small 0.527 & \small 0.177 & \small 0.047 & \small 0.303     \\
\small 4 & \small 0.529 & \small 0.267 & \small 0.154 & \small 0.170 & \small 0.572 & \small 0.201 & \small 0.040 & \small 0.332     \\
\small 5 & \small 0.631 & \small 0.262 & \small 0.173 & \small 0.195 & \small 0.581 & \small 0.175 & \small 0.060 & \small 0.346     \\
\small 6 & \small 0.581 & \small 0.252 & \small 0.077 & \small 0.252 & \small 0.525 & \small 0.191 & \small 0.041 & \small 0.293     \\
\small 7 & \small 0.610 & \small 0.260 & \small 0.155 & \small 0.196 & \small 0.650 & \small 0.209 & \small 0.064 & \small 0.377     \\
\small 8 & \small 0.532 & \small 0.259 & \small 0.137 & \small 0.137 & \small 0.486 & \small 0.200 & \small 0.048 & \small 0.238     \\
\small 9 & \small 0.571 & \small 0.274 & \small 0.180 & \small 0.118 & \small 0.514 &\small  0.229 & \small 0.084 & \small 0.202    \\
\small 10 & \small 0.645 & \small 0.306 & \small 0.087 & \small 0.252 & \small 0.612 & \small 0.202     & \small 0.054     & \small 0.356     \\ \midrule
\small Average & \small 0.573 & \small 0.259 & \small 0.132 & \small 0.183     & \small 0.542 & \small 0.195 & \small 0.052 & \small 0.294     \\
\bottomrule
\end{tabular}
\vspace*{-2mm}
\caption{WER, substitution rate, deletion rate, and insertion rate by group.}
\vspace*{-2mm}
\label{table_wer}
\end{table*}

We also noticed Whisper would insert words when there was no speech recognized in the audio clip. Typically words that had been previously transcribed would be repeated during void sections, and only one word or phrase would be repeated. This word or phrase would also be repeated for each second during the break, which created a large amount of insertions and threw off some data while testing. Finally, we found Google ASR would also hear words incorrectly and misinterpret what the speaker was saying, replacing the intended words with homonyms or phonological near-neighbors. 

\paragraph{Difference in resulting annotation labels}

The difference in labels when going from automatic segments to oracle segments can be significant. A particular case is the annotation of interruptions, one of the CPS indicators in question. Relying on the automatic segments only may split or lump utterances separated by an interruption, which may cause annotators to miss the interruption entirely (because they are only coding utterance by automatically segmented utterance), or lumping an ``interruption'' annotation with annotations of other meaningful indicators in a single, multi-speaker ``utterance.'' For tasks like this, each utterance is important as it can be the one where the correct solution has been proposed. Interestingly, \citet{bradford2023automatic} found that prosodic features were essential for identifying interruptions when using automatically segmented utterances, indicating there may be times automatic segmentation methods capture information not present in oracle contexts. 


\begin{figure}[h!]
    \centering
    \vspace*{-2mm}
    \includegraphics[width=.48\textwidth, clip]{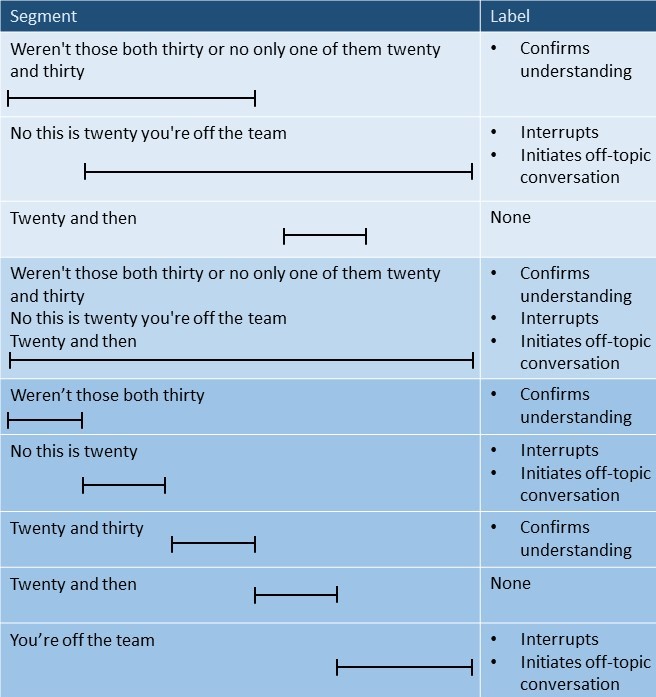}
    \vspace*{-4mm}
    \caption{Overlap between oracle (top), Google (middle), and Whisper (bottom) segments. Right column shows the CPS indicator annotated for each utterance.}
    \label{fig:Oracle-auto_overlap}
    \vspace*{-2mm}
\end{figure}

One example of label difference can be seen in the utterances shown in Fig.~\ref{fig:Oracle-auto_overlap}, with the different segmentations provided by oracle and automatic segmentation. The utterances ``Weren't those both thirty or no only one of them twenty and thirty", ``No this is twenty you're off the team" and ``Twenty and then" (which were each spoken by a different person), are combined into one segment by Google voice activity detection (VAD). The first utterance should have the label \textit{confirms understanding}, the second utterance should have the labels \textit{interrupts} and \textit{initiates off-topic conversation}, and the third utterance should have no label. However, when these are combined, all of the labels are inherited and the distinction between the different content supplied by each utterance is lost. Whisper segments split up continuous utterances by a single person, and thus person 2's {\it interrupts} and {\it initiates off-topic conversation} indicators are applied to two separate segments.  Both of these cases can cause confusion in downstream semantic classification tasks like classifying CPS indicators from linguistic features, as in~\citet{bradford2023automatic}, if the target label for training is not clear.

In some instances, the participant pauses mid-sentence and the true utterance gets split into two, but for lack of context only one gets assigned a CPS indicator. For example, a participant says ``Think it just feels like it's," pauses for 0.3 seconds, and says, ``A lot heavier because it's denser and like just carrying that.", the whole sentence should get coded as {\it discussing results}, but since the automatic segmentation splits the sentence into two utterances, only the second utterance would be coded as such.


\vspace*{-2mm}
\section{Discussion, Recommendations, and Conclusion}
\vspace*{-2mm}

One important point to emphasize is the manual cost of annotating for collaborative problem solving (CPS). CPS is a difficult annotation scheme to master (training can last as long as 6 months, depending on how much time a coder is putting toward learning). Although this paper largely focuses on the automated processes of segmenting audio, annotations themselves require complete multimodal context including viewing of video, listening to intonation, and the inclusion of temporal context. If these annotations, performed with access to multimodal information, are subsequently applied to automatically-segmented audio, then the information lost can be expected to impact downstream tasks trained or evaluated over the annotated data, thus potentially wasting the time taken to train annotators properly.

While automatic segmentation of utterances for various semantic annotation tasks certainly saves time and annotator effort, it comes at a potentially significant cost to the quality of annotations for downstream tasks. Particularly, automatic segmentation and transcription methods certainly segment utterances differently from a human oracle transcriber, and different ASR methods perform segmentation drastically differently with profoundly divergent results.  This may result in utterances being missed by the automatic segmenter or invented out of whole cloth, which would cause annotators annotating at the automatically-segmented utterance level to likewise omit annotations, or to encounter ``hallucinated'' segments that are either un-annotatable or, if annotated, introduce semantic noise into the data.  Beyond the obvious, we have shown that annotating at the oracle utterance level but then transferring those utterances to the automatically-segmented utterance level may obscure the semantic information originally captured at the oracle utterance level. Even taking the more labor-intensive step of generating oracle transcriptions before annotating is less useful if annotation is not performed at the same level. This backs up previous conclusions in semantic annotation over text-only corpora, such as the need for annotators to come to consensus on both spans and annotations~\cite{pustejovsky2012natural}, and shows that they also apply to multimodal use cases.

However, as multimodal AI develops and becomes more integrated with everyday life, inference will necessarily be performed over automatically segmented and transcribed inputs. Therefore, future models will benefit from annotation specs themselves that are task-aware and can take into account potential noise introduced by imperfect automated transcription and adjust accordingly.  For instance, if multiple labels are not allowed, should certain labels ``dominate'' others in case multiple labels are squeezed into the same segment? Future semantic annotation schemes, specifications, and languages, particularly over multimodal data, will need to take into account these requirements to more effectively use automated techniques like ASR as part of larger annotation and inference pipelines. 

\section*{Acknowledgments}

We would like to thank our reviewers for their helpful comments.  This work was supported in part by the United States National Science Foundation (NSF) under grant number DRL 2019805 to Colorado State University. The views expressed are those of the authors and do not reflect the official policy or position of the U.S. Government. All errors and mistakes are, of course, the responsibilities of the authors.

\bibliographystyle{acl_natbib}
\bibliography{anthology,acl2021}


\end{document}